\documentclass[runningheads]{llncs}
\usepackage{graphicx}
\usepackage{amsmath,amssymb} 
\usepackage{color}
\usepackage{mathtools}
\usepackage{enumerate}
\usepackage {dsfont}
\usepackage{booktabs}
\usepackage{times}
\usepackage{epsfig}
\usepackage{balance}
\usepackage{textcomp}
 \usepackage[width=122mm,left=12mm,paperwidth=146mm,height=193mm,top=12mm,paperheight=217mm]{geometry}
 
\begin{document}
\pagestyle{headings}
\mainmatter
\title{HOPC: Histogram of Oriented Principal Components of 3D Pointclouds for Action Recognition} 

\titlerunning{HOPC: Histogram of Oriented Principal Components for Action Recognition}

\authorrunning{H. Rahmani \and A. Mahmood \and D.Q. Huynh \and A. Mian}

\author{Hossein Rahmani \and Arif Mahmood \and Du Q Huynh \and Ajmal Mian}
\institute{Computer Science and Software Engineering, The University of Western Australia, \\35 Stirling Highway, Crawley, WA 6009 Australia \\
}

\maketitle

\begin{abstract}
Existing techniques for 3D action recognition are sensitive to viewpoint variations because they extract features from depth images which change significantly with viewpoint. In contrast, we directly process the pointclouds and propose a new technique for action recognition which is more robust to noise, action speed and viewpoint variations. Our technique consists of a novel descriptor and keypoint detection algorithm. The proposed descriptor is extracted at a point by encoding the Histogram of Oriented Principal Components (HOPC) within an adaptive spatio-temporal support volume around that point. Based on this descriptor, we  present a novel method to detect Spatio-Temporal Key-Points (STKPs) in 3D pointcloud sequences. Experimental results show that the proposed descriptor and STKP detector outperform state-of-the-art algorithms on three benchmark human activity datasets. We also introduce a new multiview public dataset and show the robustness of our proposed method to viewpoint variations.
\end{abstract}

\section{Introduction}
Human action recognition has many applications in smart surveillance, human-computer interaction and sports. The Kinect and other depth cameras have become popular for this task because depth sequences do not suffer from the problems induced by variations in illumination and clothing texture. However, the presence of occlusion, sensor noise and most importantly viewpoint variations still make action recognition a challenging task. 

Designing an efficient depth sequence representation is an important task in many computer vision problems. Most existing action recognition techniques (e.g.,~\cite{CCD,MyWACV14,DMM}) treat depth sequences the same way as color videos and use color-based action recognition methods. However, while these methods are suitable for color video sequences, simply extending them to depth sequences may not be optimal~\cite{HON4D}. Information captured by depth cameras actually allows geometric features to be extracted to form rich descriptors. For instance, Tang et al.~\cite{HONV} used histograms of the normal vectors for object recognition in depth images. Given a depth image, they computed spatial derivatives, transformed them to the polar coordinates and used the 2D histograms as object descriptors. Recently, Oreifej and Liu~\cite{HON4D} extended the same technique to the temporal dimension by adding time derivative. A downside of treating depth sequences this way is that the noise in the depth images is enhanced by the differential operations~\cite{Wang2012}. Histogramming, on the other hand, is analogous to integration and is more resilient to the effect of noise. Furthermore, viewpoint variations are unavoidable in real scenarios. However, none of the existing 3D sensor based techniques is designed for cross-view action recognition where training is performed on sequences acquired from one view and testing is performed on sequences acquired from a significantly different view ($>25^\circ$).

\begin{figure}[t]
\begin{center}
\includegraphics[width=11 cm]{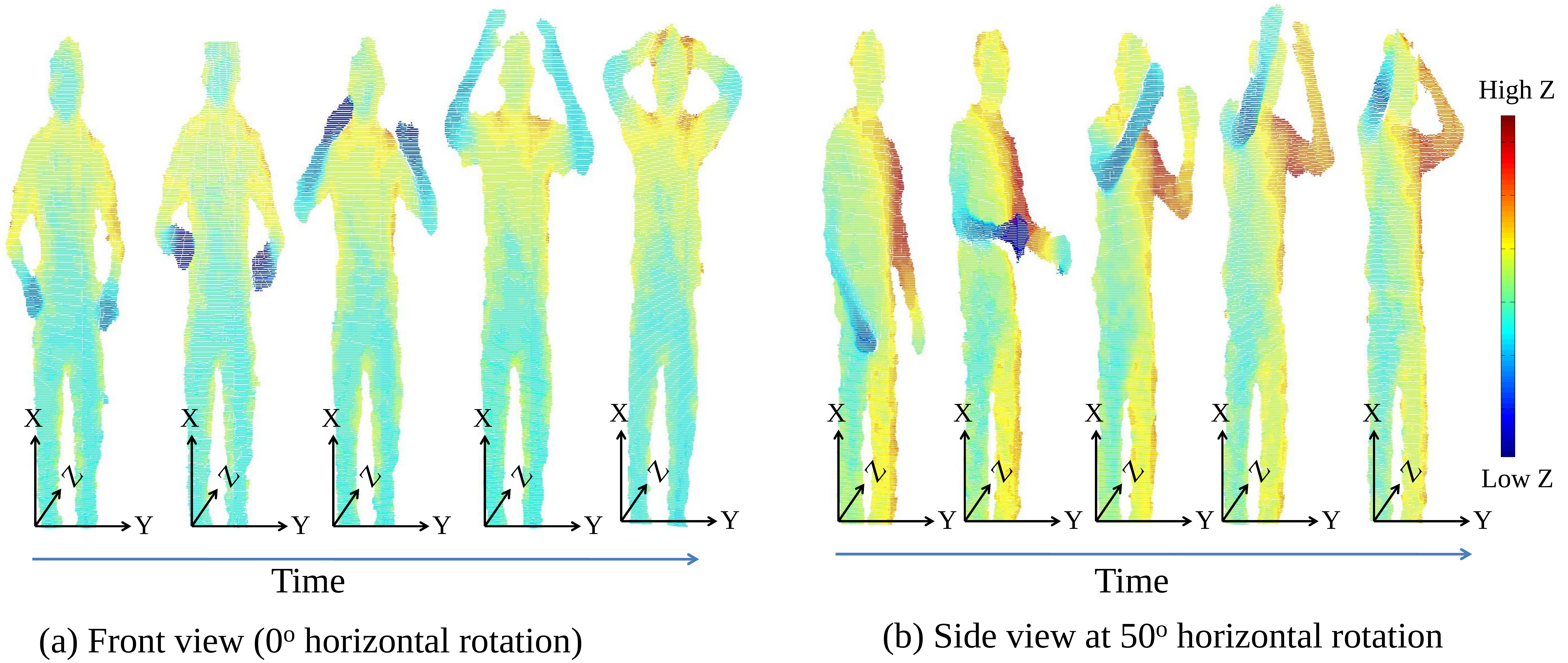}
\end{center}
\vspace{-6mm}
\caption{Two sequences of 3D pointclouds of a subject performing the {\it holding head} action. Notice how the depth values (colours) have significantly changed with the change in viewpoint. Simple normalization cannot compensate for such depth variations. Existing  depth based action recognition algorithms will not be accurate in such cases}
\label{fig:PCloudSeq}
\end{figure}

We directly process the 3D pointcloud sequences (Fig.~\ref{fig:PCloudSeq}) and extract point descriptors which are robust to noise and viewpoint variations. We propose a novel descriptor, the {\it Histogram of Oriented Principal Components} (HOPC), to capture the local geometric characteristics around each point within a sequence of 3D pointclouds. To extract HOPC at a point ${\bf p}$, PCA is performed on an adaptive spatio-temporal support volume around ${\bf p}$ (see Fig.~\ref{fig:KeyPointAlg}) which gives us a $3 \times 3$ matrix of eigenvectors and the corresponding eigenvalues. Each eigenvector is projected onto $m$ directions corresponding to the vertices of a {\it regular m-sided polyhedron} and scaled by its eigenvalue. HOPC is formed by concatenating the projected eigenvectors in decreasing order of their eigenvalues. 

HOPC is used in a holistic and local setting. In the former approach, the sequence of 3D pointclouds is divided into spatio-temporal cells and HOPC descriptors of all points within a cell are accumulated and normalized to form a single cell descriptor. All cell descriptors are concatenated to form a holistic HOPC descriptor. In the latter approach, local HOPC are extracted at candidate spatio-temporal keypoints (STKP) and a HOPC quality factor is defined to rank the STKPs. Only high quality STKPs are retained. All points within the adaptive spatio-temporal support volume of each STKP are aligned along the eigenvectors of the spatial support around STKP. Thus the support volume is aligned with a local object centered coordinate basis and extracting HOPC, or any other feature, at the STKP will be view invariant. See Section \ref{Local} for details. Since humans may perform the same action at different speeds, to achieve speed invariance, we propose automatic temporal scale selection by minimizing the eigenratios over a varying temporal window size. The main contributions of this paper include:
\begin{itemize}
  \item A HOPC descriptor for 3D pointclouds.
  \item A spatio-temporal key-point (STKP) detector and a view invariant descriptor.
  \item A technique for speed normalization of actions.
  \end{itemize}  

Moreover, we introduce a new 3D action dataset which has scale variations of subjects and viewpoint variations. It contains thirty actions which is larger number than any existing 3D action dataset. This dataset will be made public. 
Experimental comparison on four datasets, including three benchmark ones \cite{Bag3DPoints,HON4D,ActionLet2012},  with eight state-of-the-art methods \cite{CCD,3Dgrad,HON4D,MyWACV14,Wang2012,ActionLet2012,DSTIP,ViewInvariantJoint3D} shows the efficacy of our algorithms. Data and code of our technique are available \cite{HOPC_code}.
\section{Related Work}
Based on the input data, human action recognition methods can be divided into three categories including RGB based,  skeleton-based and depth based methods. In RGB videos, in order to recognize actions across viewpoint changes, mostly view independent representations are proposed such as view invariant spatio-temporal features \cite{3,17,21,22,23,28}. Some methods infer the 3D scene structure and use geometric transformations to achieve view invariance \cite{4,10,15,26,29}. Another approach is to find a view independent latent space \cite{8,7,virtualviews,14} in which features extracted from the actions captured at different view points are directly comparable. Our proposed approach also falls in this category. However, our approach is only for 3D pointclouds captured by depth sensors. To the best of our knowledge, we are the first to propose cross-view action recognition using 3D pointclouds. We propose to normalize the spatio-temporal support volume of each candidate keypoint in the 3D pointcloud such that the feature extracted from the normalized support volume becomes view independent. 
 
In skeleton based methods, 3D joint positions are used for action recognition. Multi-camera motion capture
(MoCap) systems \cite{mocap} have been used for human action
recognition, but such special equipment is marker-based
and expensive. Moreover, due to the different quality of
the motion data, action recognition methods designed
for MoCap  are not suitable for 3D pointcloud sequences which is the focus of this paper \cite{ActionLet2012}. 

On the other hand, some methods~\cite {ViewInvariantJoint3D,Wang2012,eigenjoints} use the human joint positions extracted by the OpenNI tracking framework (OpenNI)~\cite{SingleDepth} as interest points. For example, Yang and Tian~\cite{eigenjoints} proposed pairwise 3D joint position differences in each frame and temporal differences across frames to represent an action. Since 3D joints cannot capture all the discriminative information, the action recognition accuracy is compromised. Wang et al.\ \cite{ActionLet2012} extended the previous approach by computing the histogram of occupancy pattern of a fixed region around each joint in a frame. In the temporal dimension, they used low frequency Fourier components as features and an SVM to find a discriminative set of joints. It is important to note that the estimated joint positions are not reliable and can fail when the human subject is not in an upright and frontal view position (e.g. lying on sofa) or when there is clutter around the subject.

Action recognition methods based on depth maps can be divided into holistic \cite{HON4D,MyWACV14,Bag3DPoints,DMM,STOP} and local approaches~\cite{ActionLet2012,DSTIP,STIP,Wang2012}. Holistic methods use global features such as silhouettes and space-time volume information. For example, Li et al.\ \cite{Bag3DPoints} sampled boundary pixels from 2D silhouettes as a bag of features. Yang et al.\ \cite{DMM} added temporal 
derivative of 2D projections to get Depth Motion Maps (DMM). Vieira et al.\ \cite{STOP} computed silhouettes in 3D by using the space-time occupancy patterns. Recently, Oreifej and Liu~\cite{HON4D} extended histogram of oriented 3D normals \cite{HONV} to  4D by adding time derivative. The gradient vector was normalized to unit magnitude and projected on a refined basis of  600-cell Polychrome to make histograms. The last component of normalized gradient vector was inverse of the gradient magnitude. As a result, information from very strong derivative locations, such as edges and silhouettes, may get suppressed~\cite{MyWACV14}. The proposed HOPC descriptor is more informative than HON4D as it captures the spread of data in three principal directions. Thus, HOPC achieves more action recognition accuracy than exiting methods on three benchmark datasets. 

Depth based local methods use local features where a set of interest points are extracted from the depth sequence and a feature descriptor is computed for each interest point. For example, Cheng et al.~\cite{CCD} used interest point detector proposed by Doll\'{a}r et al.~\cite{STIP} and proposed a Comparative Coding Descriptor (CCD). Due to  the presence of noise in depth sequences, simply extending color-based interest point detectors such as~\cite{Dollar} and~\cite{STIP} may degrade the efficiency of these detectors~\cite{HON4D}. 

Motion trajectory based action recognition methods\cite{DensTraj,Traj} are also not reliable in depth sequences~\cite{HON4D}. Therefore, recent depth based action recognition methods resorted to alternative ways to extract more reliable interest points. Wang et al. \cite{Wang2012} proposed Haar features to be extracted from each random subvolume.  Xia and Aggarwal in~\cite{DSTIP} proposed a filtering method to extract spatio-temporal interest points. Their approach fails when the action execution speed is faster than the flip of the signal caused by the sensor noise. Both techniques are sensitive to viewpoint variations. 

In contrast to previous interest point detection methods, the proposed STKP detector is robust to variations in action execution speed, sensor viewpoint and the spatial scale of the actor. Since the proposed HOPC descriptor is not strictly based on the depth derivatives,  it  is more robust to noise. Moreover, our methods do not require skeleton data which may be noisy or unavailable especially in the case of side views.
\section{Histogram of Oriented Principal Component (HOPC)}
\label{HOPCDescriptor}
Let $Q=\{Q_1, Q_2,\cdots,Q_t,\cdots,Q_{n_f}\}$ represent a sequence of 3D pointclouds captured by a 3D sensor, where $n_f$ denotes the number of frames (i.e. number of 3D pointclouds in the sequence) and $Q_t$ is the 3D pointcloud at time $t$. We make a spatio-temporal accumulated 3D pointcloud by merging the sequence of individual pointclouds in the time interval $[t-\tau,t+\tau]$. Consider a point ${\bf p} = (x_t \; y_t \; z_t)^{\top}, 1 \leq t \leq n_f$ in $Q_t$. We define the spatio-temporal support of ${\bf p}$, $\Omega({\bf p})$, as the 3D points which are in a sphere of radius $r$ centered at ${\bf p}$ (Fig. \ref{fig:KeyPointAlg}). We propose a point descriptor based on the eigenvalue decomposition of the scatter matrix $C$ of the points ${{\bf q} \in \Omega({{\bf p}})}$:
\begin{equation} C = \frac{1}{n_p} \sum_{{\bf q} \in \Omega({{\bf p}})}
{({\bf q} - {\mu}){({\bf q} - {\mu})}^{\top}}, \text{where}\; \mu = \frac{1}{n_p} \sum_
{{\bf q} \in \Omega({{\bf p}})}{\bf q},\label{eq:ScatterMatrix} \end{equation} 
and $n_p=|\Omega({{\bf p}})|$ denotes the number of points in the spatio-temporal support of ${\bf p}$. Performing PCA on the scatter matrix $C$ gives us $CV = EV$, where $E$ is a diagonal matrix of the eigenvalues $\lambda_1 \geq \lambda_2 \geq \lambda_3$, and $V$ contains three orthogonal eigenvectors $[{\bf v}_1 \; {\bf v}_2 \; {\bf v}_3]$ arranged in the order of decreasing magnitude of their associated eigenvalues. 
We propose a new descriptor, the Histogram of Oriented Principal Components (HOPC), by projecting each eigenvector onto $m$ directions obtained from a {\it regular m-sided polyhedron}. We use $m=20$ to make a {\it regular icosahedron} which is composed of $20$ {\it regular pentagonal} facets and each facet corresponds to a histogram bin. Let $U \in \mathds{R}^{3 \!\times\! m}$ be the matrix of the center positions ${\bold u}_1, {\bold u}_2, \cdots , {\bold u}_{m}$ of facets:
\begin{equation}U=[{\bold u}_1,{\bold u}_2,\cdots,{\bf u}_i,\cdots,{\bold u}_{m}]\end{equation} For a {\it regular icosahedron} with center at the origin, these normalized vectors are
\begin{equation}
\left(\frac{\pm1}{L_u},\frac{\pm1}{L_u},\frac{\pm1}{L_u}\right),\left(0,\frac{\pm\varphi^{-1}}{L_u},\frac{\pm\varphi}{L_u}\right),\left(\frac{\pm\varphi^{-1}}{L_u},\frac{\pm\varphi}{L_u},0\right),\left(\frac{\pm\varphi}{L_u},0,\frac{\pm\varphi^{-1}}{L_u}\right),
\end{equation}
where $\varphi={(1+\sqrt{5})}/{2}$ is the golden ratio, and $L_u=\sqrt{\varphi^2+{1}{/\varphi^2}}$ is the length of vector ${\bf u}_i, 1 \leq i \leq m$. The eigenvectors are basically directions of maximum variance of the points in 3D space. Thus, they have a $180^\circ$ ambiguity. To overcome this problem, we consider the distribution of vector directions and their magnitudes within the support volume of ${\bf p}$. We determine the sign of each eigenvector ${\bf v}_j$ from the sign of the inner products of ${\bf v}_j$ and all vectors within the support of ${\bf p}$:
\begin{equation}
{\bf v}_j={\bf v}_j.sign\left(\sum_{{\bf q} \in \Omega({\bf p})}{sign({\bf o}^{\top}{\bf v}_j)({\bf o}^{\top}{\bf v}_j})^2 \right)
\end{equation}
where ${\bf o}={\bf q}-{\bf p}$ and the $sign$ function returns the sign of an input number. Note that the squared projection ensures the suppression of small projections, which could be due to noise. If the signs of eigenvectors ${\bf v}_1, {\bf v}_2,$ and ${\bf v}_3$ disagree i.e. ${\bf v}_1 \times {\bf v}_2 \neq {\bf v}_3$, we switch the sign of the eigenvector whose $|\sum_{w=1}^{n_p}{sign({\bf o}_w^{\top}{\bf v}_j)({\bf o}_w{\bf v}_j})^2|$ value is the smallest. We then project each eigenvector ${\bf v}_j$ onto $U$ to give us:
\begin{equation}{\bf b}_j=U^{\top} {\bf v}_j \in \mathds{R}^{m}, \textbf{ for } 1 \le j \le 3. \end{equation} In case ${\bf v}_j$ is perfectly aligned with ${\bf u}_i \in U$, it should vote into only $i^{\text{th}}$ bin. However, all ${\bf u}_i$'s are not orthogonal, therefore ${\bf b}_j$ will have non-zero projection in other bins as well. To overcome this effect, we quantize the projection of ${\bf b}_j$. For this purpose, a threshold value $\psi$ is computed by projecting any two {\it neighbouring} vectors ${\bf u}_k$ and ${\bf u}_l$,
\begin{equation}\psi={{\bf u}_k}^{\top} {\bf u}_l=\frac{\varphi+\varphi^{-1}}{{L_u}^2},\;\;{\bf u}_k,{\bf u}_l \in U.\end{equation} Note that for any ${\bf u}_k \in U$, we can find a ${\bf u}_l \in U$ such that $\psi=({\varphi+\varphi^{-1}})/{L_u}^2$. The quantized vector is given by
\[ \hat{{\bf b}}_j(z) = \left\{
  \begin{array}{l l}
    0 & \quad \text{if ${\bf b}_j(z) \le \psi$}\\
    {\bf b}_j(z)-\psi & \quad \text{otherwise},
  \end{array} \right.\]
where $1 \leq z \leq m$. We define ${\bf h}_j$ to be $\hat{{\bf b}}_j$ scaled by the corresponding eigenvalue $\lambda_j$,
\begin{equation}  {\bf h}_j=\frac {\lambda_j \cdot \hat{{\bf b}}_j}
  {||\hat{{\bf b}}_j||_2} \in \mathds{R}^{m}, \textbf{ for } 1 \le j \le 3.
\end{equation} We concatenate the histograms of oriented principal components of all three eigenvectors in decreasing order of their eigenvalues to form a descriptor of point ${\bf p}$:
\begin{equation} \label{HOPC} {\bf h}_{\bf p}=[{\bf h}_1^{\top}\; {\bf h}_2^{\top}\; {\bf h}_3^{\top}]^{\top} \in \mathds{R}^{3m}. \end{equation}

The spatio-temporal HOPC descriptor at point ${\bf p}$ encodes information from both shape and motion in the support volume around it. Since the smallest principal component of the local surface is in fact the total least squares estimate of the surface normal~\cite{surfaceNormal}, our descriptor, which inherently encodes the surface normal, is more robust to noise than gradient-based surface normal used in~\cite{HONV,HON4D}. Using this descriptor, we propose two different action recognition algorithms in the following section.
\section{Action Recognition}
We propose a holistic and a local approach for human action recognition. Our holistic method is suitable for actions under occlusions, more inter-class similarities of local motions, and where the subjects do not change their spatial locations. On the other hand, our local method is more suitable for cross-view action recognition and in cases where the subjects change their spatial locations.
\subsection{Action Recognition with Holistic HOPC}
\label{Holistic}
A sequence of 3D pointclouds is divided into $\gamma= n_x \times n_y \times n_t$ spatio-temporal cells along $X$, $Y$, and $T$ dimensions. We use $c_s, \text{where}\; s=1 \cdots \gamma$, to denote the $s^{\text{th}}$ cell. The spatio-temporal HOPC descriptor ${\bf h}_{\bf p}$ in \eqref{HOPC} is computed for each point ${\bf p}$ within the sequence. The cell descriptor ${\bf h}_{c_s}$ is computed by accumulating ${\bf h}_{c_s}=\sum_{p \in c_s}{{\bf h}_{\bf p}}$ and then normalizing ${\bf h}_{c_s}\leftarrow{{\bf h}_{c_s}}/{||{\bf h}_{c_s}||_2}$. The final descriptor ${\bf h}_{v}$ for the given sequence is a concatenation of ${\bf h}_{c_s}$ obtained from all the cells:${\bf h}_{v}={[{\bf h}_{c_1}^{\top}\; {\bf h}_{c_2}^{\top}\; ...\; {\bf h}_{c_s}^{\top}\; ...\; {\bf h}_{c_{\gamma}}^{\top}]}^{\top}$. We use ${\bf h}_{v}$ as the holistic HOPC descriptor and use SVM for classification.
\subsubsection{Computing a Discriminative Cell Descriptor:}
The HOPC descriptor is highly correlated to the order of eigenvalues of the spatio-temporal support volume around ${\bf p}$. Therefore, for each point a pruning approach is introduced to eliminate the ambiguous eigenvectors of each point. For this purpose, we define two eigenratios:
\begin{equation}\delta_{12}=\frac{\lambda_1}{\lambda_2}, 
\delta_{23}=\frac{\lambda_2}{\lambda_3}.\end{equation}

For 3D symmetrical surfaces, the values of $\delta_{12}$ or $\delta_{23}$ will be equal to $1$. The principal components of symmetrical surfaces are ambiguous. To get a discriminative ${\bf h}_{\bf p}$, the values of $\delta_{12}$ and $\delta_{23}$ must be greater than 1. However, to manage noise we choose a threshold value $\theta > 1+\epsilon$, where $\epsilon$ is a margin and select only the discriminative eigenvectors as follows:
\begin{enumerate}
\item If $\delta_{12} > \theta$ and $\delta_{23} > \theta$: ${\bf
  h}_{\bf p}=[{\bf h}_1^{\top}\;\;{\bf h}_2^{\top}\;\; {\bf
    h}_3^{\top}]^\top$.\vspace{+1mm}
\item If $\delta_{12} \le \theta$ and $\delta_{23} > \theta$:
  ${\bf h}_{\bf p}=[{\bf 0}^{\top}\;\;{\bf 0}^{\top}\;\; {\bf
    h}_3^{\top}]^\top$.\vspace{+1mm}
\item If $\delta_{12} > \theta$ and $\delta_{23} \le \theta$:
  ${\bf h}_{\bf p}=[{\bf h}_1^{\top}\;\;{\bf 0}^{\top}\;\; {\bf
    0}^{\top}]^\top$.\vspace{+1mm}
\item If $\delta_{12} \le \theta$ and $\delta_{23} \le \theta$: In this case,
  we discard ${\bf p}$.
\end{enumerate}
\subsection{STKP: Spatio-Temporal Key-Point Detection}
\label{Local}
Consider a point ${\bf p}=(x_t\; y_t\; z_t)^{\top}$ within a sequence of 3D pointclouds. In addition to the spatio-temporal support volume around ${\bf p}$ defined in section~\ref{HOPCDescriptor}, we further define a spatial only support volume around ${\bf p}$ as the 3D points of $Q_t$ that fall inside  a sphere of radius $r$ centered at ${\bf p}$. Thus, we perform PCA on both the spatial and the spatio-temporal scatter matrices $C'$ and $C$. 

Let $\lambda_1' \geq \lambda_2' \geq \lambda_3'$ and $\lambda_1 \geq \lambda_2 \geq \lambda_3$ represent the eigenvalues of the spatial $C'$ and spatio-temporal $C$ scatter matrix, respectively. We define the following ratios:
\begin{equation}
\delta_{12}'=\frac{\lambda_1'}{\lambda_2'},\; 
\delta_{23}'=\frac{\lambda_2'}{\lambda_3'},\;
\delta_{12}=\frac{\lambda_1}{\lambda_2},\; 
\delta_{23}=\frac{\lambda_2}{\lambda_3}.\;
\end{equation}

For a point to be identified as a potential keypoint, the condition $\{\delta_{12},\delta_{23},\delta_{12}',\delta_{23}'\} > \theta$ must be satisfied. This process prunes ambiguous points and produces a subset of candidate keypoints. It reduces the computational burden of the subsequent steps. Let ${\bf h}'_{\bf p} \in \mathds{R}^{3m}$ represent the spatial HOPC and ${\bf h}_{\bf p} \in \mathds{R}^{3m}$ represent the spatio-temporal HOPC. A {\it quality} factor is computed at each candidate keypoint ${\bf p}$ as follows:
\begin{equation}
\eta_p=\frac{1}{2}\sum_{i=1}^{3m}{\frac{({\bf h}'_{\bf p}(i)-{\bf h}_{\bf p}(i))^2}{({\bf h}'_{\bf p}(i)+{\bf h}_{\bf p}(i))}}.
\end{equation}

\noindent When ${\bf h}_{\bf p}'={\bf h}_{\bf p}$, the {\it quality} factor has the minimum value of $\eta_p=0$. It means that the candidate keypoint ${\bf p}$ has a stationary spatio-temporal support volume with no motion. 

\begin{figure}[t]
\begin{center}
\includegraphics[width=7.5 cm]{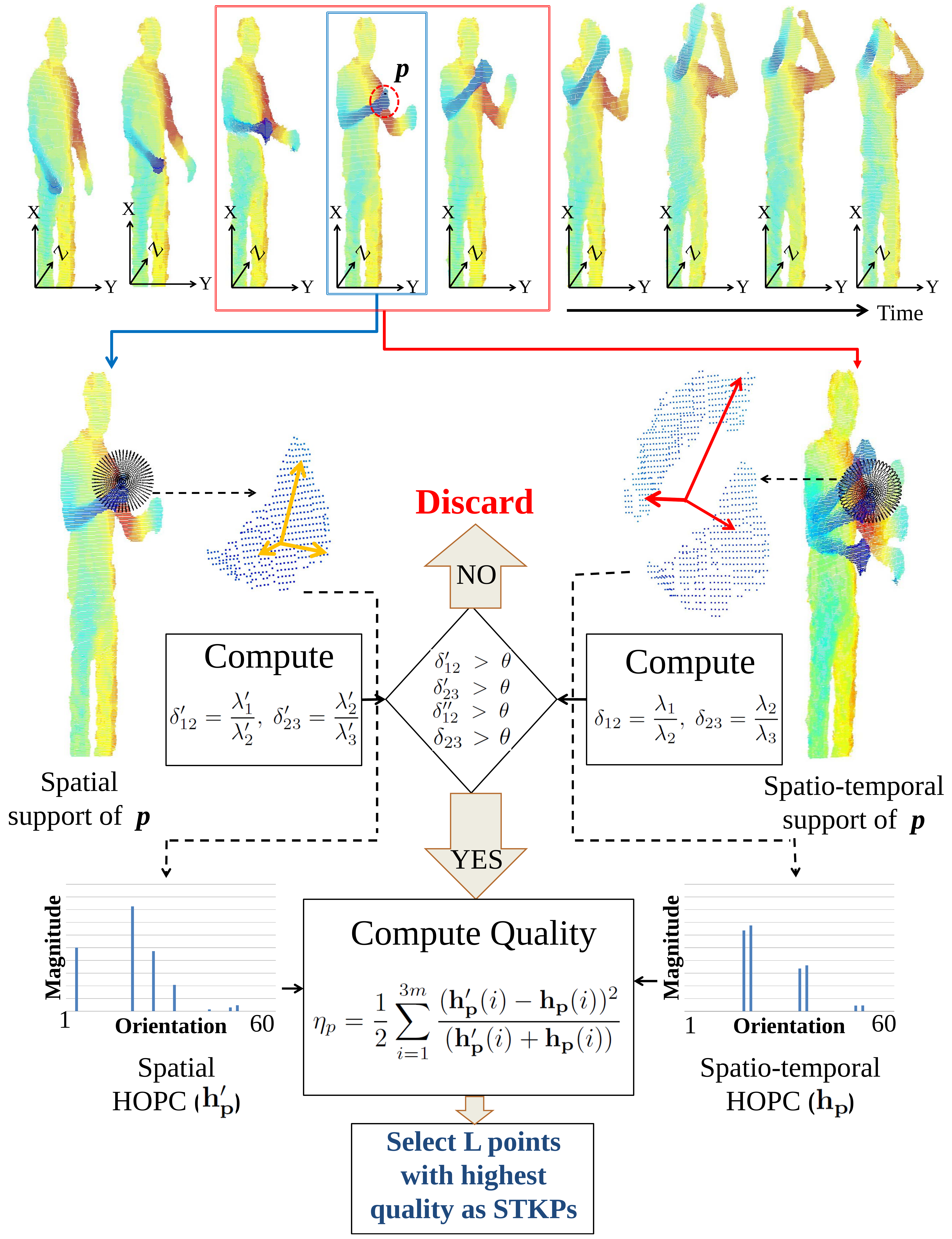}
\end{center}
\vspace{-6mm}
\caption{STKP: Spatio-Temporal Key-Point detection algorithm}
\label{fig:KeyPointAlg}
\end{figure}

\begin{figure}[t]
\begin{center}
\includegraphics[width=11 cm]{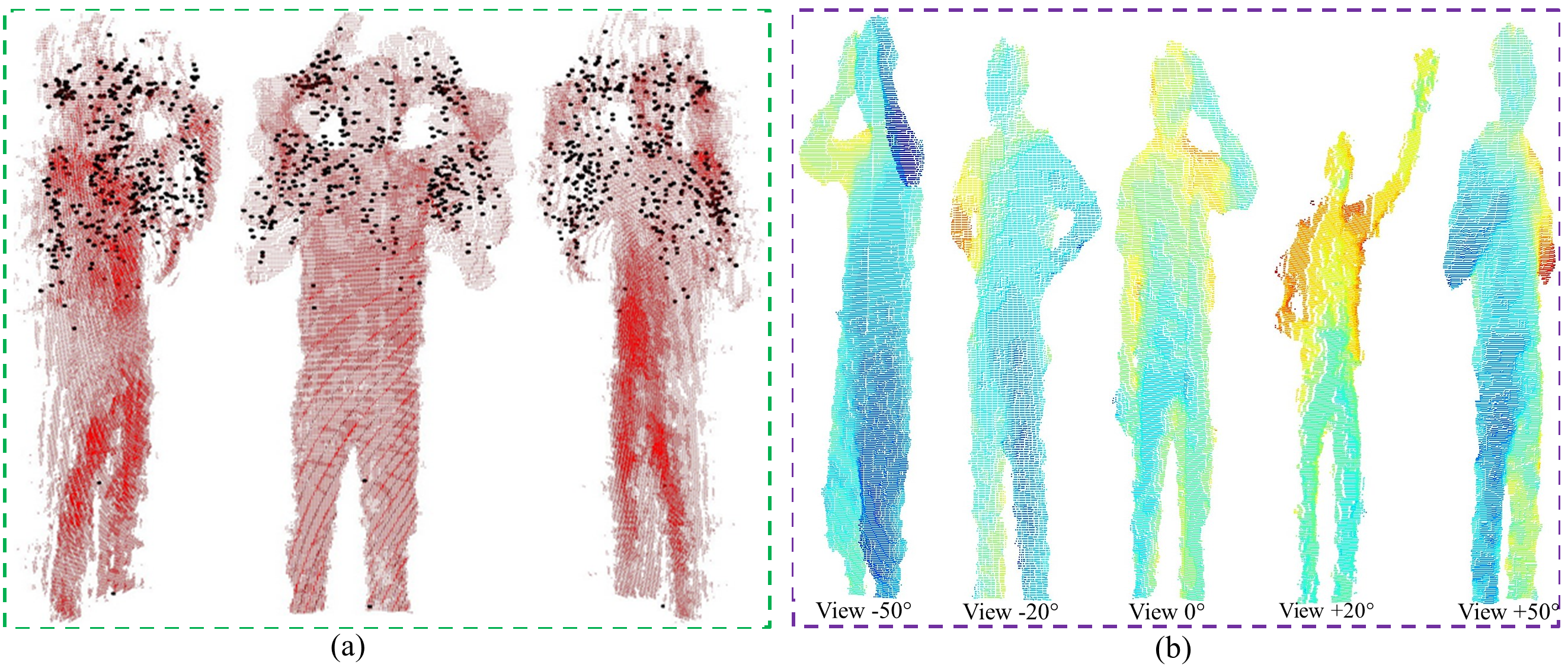}
\end{center}
\vspace{-6mm}
\caption{(a)-STKPs projected onto $XY$ dimensions on top of all points within a sequence of 3D pointclouds corresponding to the {\it holding head} action (from three different views). Note that a large number of STKPs are detected only where movement is performed. (b)-Sample pointclouds at different views from the UWA3D Multiview Activity dataset}
\label{fig:Keypoints}
\end{figure}

We define a locality as a sphere of radius $r'$ ( with $r'\ll r$) and a time interval $2\tau'+1$ (with $\tau' \le \tau$). We sort the candidate STKPs according to their quality values and starting from the highest quality keypoint, all STKPs within its locality are removed. The same process is repeated on the remaining STKPs. Fig.~\ref{fig:KeyPointAlg} shows the steps of our STKP detection algorithm. Fig.~\ref{fig:Keypoints}-a shows the extracted STKPs from three different views for a sequence of 3D pointclouds corresponding to the {\it holding head} action. 
\subsection{View-Invariant Key-Point Descriptor}
Let ${\bf p}=(x_t\;y_t\;z_t)^{\top}$ represent an STKP. All points within the spatio-temporal support volume of ${\bf p}$ i.e., $\Omega({\bf p})$, are aligned along the eigenvectors of its spatial scatter matrix, $B=PV'$, where $P \in \mathds{R}^{n_p \!\times\! 3}$ is a matrix  of points within $\Omega({\bf p})$ and $V'=[{\bf v}_1' \;{\bf v}_2' \;{\bf v}_3']$ denotes the $3 \times 3$ matrix of eigenvectors of the spatial scatter matrix $C'$. Recall that the signs of these eigenvectors have a $180^\circ$ ambiguity. As mentioned earlier, we use the sign disambiguation method to overcome this problem. As a result, any feature (e.g. raw depth values or HOPC) extracted from the aligned spatio-temporal support volume around ${\bf p}$ will be view invariant. 

In order to describe the points within the spatio-temporal support volume of keypoint ${\bf p}$, the spatio-temporal support of ${\bf p}$ is represented as a 3D hyper-surface in the 4D space $(X,Y,Z)$ and $T$. We fit a 3D hyper-surface to the aligned points within the spatio-temporal support volume of ${\bf p}$. A uniform $m_x \times m_y \times m_t$ grid is used to sample the hyper-surface and its raw values are used as the descriptor of keypoint ${\bf p}$.

We use the bag-of-words approach to represent each 3D pointcloud sequence and build a codebook by clustering the keypoint descriptors using K-means. Codewords are defined by the cluster centers and  descriptors are assigned to codewords using Euclidean distance. For classification, we use SVM with the histogram intersection kernel~\cite{HIKSVM}.
\section{Adaptive Support Volume}
So far we have used a fixed spatial ($r$) and temporal ($\tau$) support volume to detect and describe each keypoint ${\bf p}$. However, subjects can have different scales (in height and width) and perform actions with different speeds. Therefore, simply using a fixed spatial ($r$) and temporal ($\tau$) support volume is not optimal. Large values of $r$ and $\tau$ enable the proposed descriptors to encapsulate more information about shape and motion of a subject. However, this also increases sensitivity to occlusion and action speed.

A simple approach to finding the optimal spatial scale ($r$) for a STKP is based on the subject's height ($h_s$) e.g.~$r=e \times h_s$, where $e$ is a constant that is empirically chosen to make a trade-off between descriptiveness and occlusion. This approach is unreliable and may fail when a subject touches the background or is not in an upright position. Several automatic spatial scale detection methods \cite{3DkeypointSurvey} have been proposed for 3D object recognition. In this paper, we use the automatic spatial scale detection method proposed by Mian et al.~\cite{Mian} to determine the optimal spatial scale for each keypoint. The optimal spatial scale  ($r_b$) is selected as the one for which the ratio between the first two eigenvalues of the spatial support of a keypoint reaches a local maximum. Our results show that the automatic spatial scale selection~\cite{Mian} achieves the same accuracy as the fixed scale when the height ($h_s$) of each subject is available.

For temporal scale selection, most previous works~\cite{HON4D,DSTIP,MyWACV14,STOP,Dollar} used a fixed number of frames. However, we propose automatic temporal scale selection to make our descriptor robust to action speed variations. Our proposed method follows the automatic spatial scale detection method by Mian et al.~\cite{Mian}. Let $Q=\{Q_1, Q_2,\cdots,Q_t,\cdots,Q_{n_f}\}$ represent a sequence of 3D pointclouds. For a point ${\bf p}=[x_t\;y_t\;z_t]^{\top}$, we start with points in $[Q_{t-\tau},\cdots,Q_{t+\tau}]$ for $\tau=1$ which are within its spatial scale $r$ (note that we assume $r$ as the optimal spatial scale for ${\bf p}$) and calculate the summation of ratio between the first two eigenvalues (${\lambda_2}/{\lambda_1}$) and the last two eigenvalues (${\lambda_3}/{\lambda_2}$) as:
\begin{equation}
A_{\bf p}^{\tau}=\frac{\lambda_2}{\lambda_1}+\frac{\lambda_3}{\lambda_2},
\end{equation}
where $\lambda_1 \geq \lambda_2 \geq \lambda_3$. This process continues for all $\tau=1,\cdots,\Delta$ and the optimal temporal scale $\tau$ corresponding to the local minimum value of $A_{\bf p}$ found for point ${\bf p}$. A point which does not have a local minimum is not considered as a candidate keypoint. 
\section{Experiments}
\label{Experiments}
The proposed algorithms  were evaluated on three benchmark datasets including MSRAction3D~\cite{Bag3DPoints}, MSRGesture3D~\cite{Wang2012}, and ActionPairs3D~\cite{HON4D}. We also developed a new ``UWA3D Multiview Activity'' dataset to evaluate the proposed cross-view action recognition algorithm. This dataset consists of 30 daily activities of ten subjects performed at different scales and viewpoints (Subsection~\ref{MyDataset}). For our algorithms, we used  $k=1000, \theta=1.12$, $m_x=m_y=20$ and $m_t=3$ in all experiments. To test the performance of our holistic approach, each sequence of 3D pointclouds was divided into $6 \times 5 \times 3$ spatio-temporal cells along $X$, $Y$, and $T$ dimensions, respectively. 

The performance of the proposed algorithms was compared with seven state-of-the-art methods including Histogram of Oriented Gradient (HOG3D)~\cite{3Dgrad}, Random Occupancy Pattern (ROP)~\cite{Wang2012}, Histogram of 3D joints(HOJ3D)~\cite{ViewInvariantJoint3D}, Actionlet Ensemble~\cite{ActionLet2012}, Histogram of 4D Oriented Normals (HON4D)~\cite{HON4D}, Depth Spatio-Temporal Interest Points (DSTIP)~\cite{DSTIP}, and Histograms of Depth Gradient (HDG)~\cite{MyWACV14}. The accuracy is reported from the original papers or from the authors' implementations of DSTIP~\cite{DSTIP}, HDG~\cite{MyWACV14}, HOG3D~\cite{3Dgrad}, and HON4D~\cite{HON4D}. The implementation of HOJ3D~\cite{ViewInvariantJoint3D} is not available, therefore we used our own implementation.

\begin{figure}[t]
\begin{center}
\includegraphics[width=12 cm]{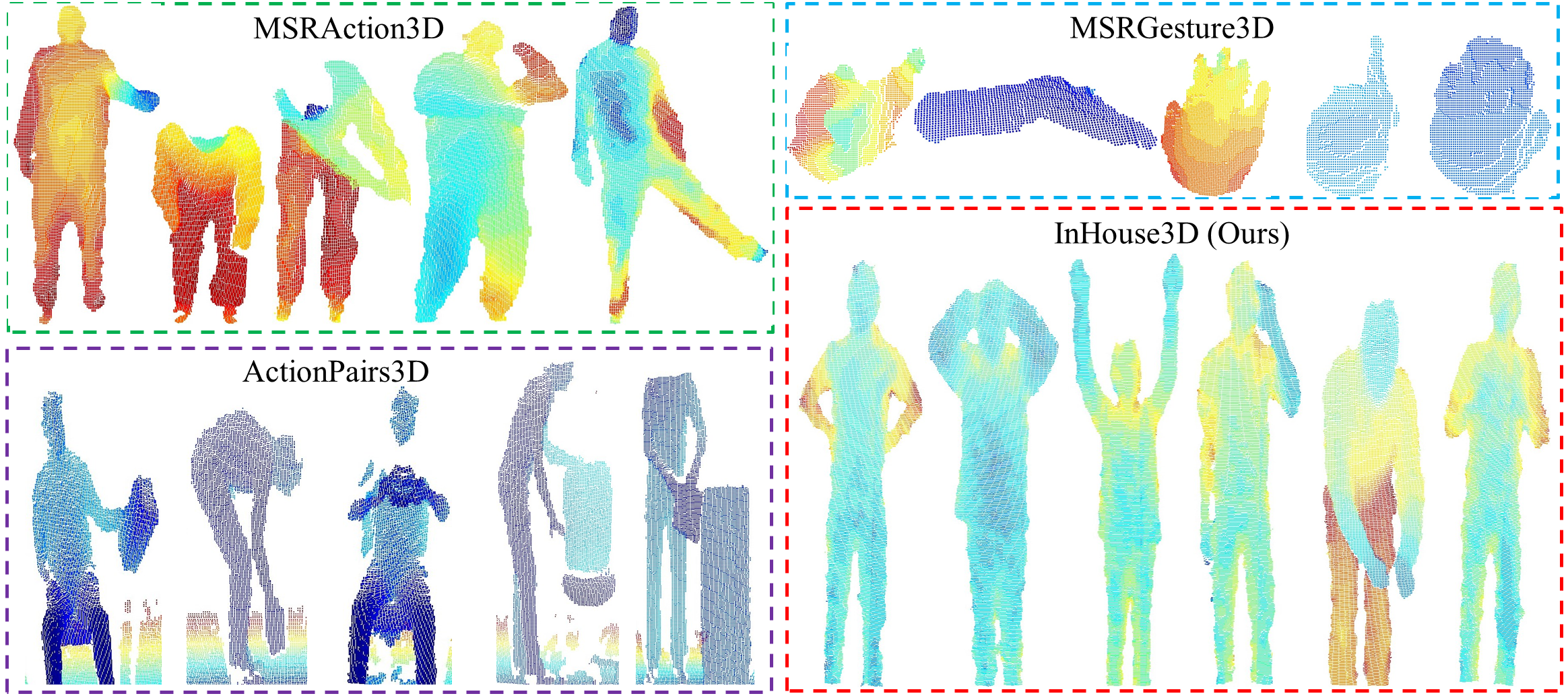}
\end{center}
\vspace{-6mm}
\caption{Sample 3D pointclouds from the MSRAction3D, MSRGesture3D, ActionPairs3D, and UWA3D Multiview Activity datasets}
\label{fig:DataSamples}
\end{figure}

\subsection{MSRAction3D dataset}
\label{MSRAction3D dataset}
MSRAction3D dataset~\cite{Bag3DPoints} consists of 20 actions each performed by 10 subjects 2-3 times (Fig.~\ref{fig:DataSamples}). The dataset is challenging due to high inter-action similarities. To test our holistic approach, we used five subjects for training and five for testing and repeated the experiments 252 folds exhaustively as proposed by~\cite{HON4D}. To show the effectiveness of our automatic spatio-temporal scale selection, we used four different settings using fixed and varying values of $r$ and $\tau$. Table \ref{tab:Action} compares our algorithms with existing state-of-the-art. Note that the proposed algorithm outperformed all techniques under all four settings. The maximum accuracy was achieved using constant $r$ and adaptive $\tau$. Adaptive $r$ did not improve results since there is little scale variation in this dataset. Note that  HOJ3D~\cite{ViewInvariantJoint3D}, Moving Pose \cite{Zanfir} and Actionlet~\cite{ActionLet2012} use skeleton data which is not always available. 

\begin{table}[t] 
 \caption{\small Accuracy comparison on MSRAction3D dataset. Mean $\pm$ STD is computed over 252 folds. Fold 5/5 means subjects \{1,3,5,7,9\} used for training and the rest for testing. $^a$ Moving Pose~\cite{Zanfir} used different setting}
 \centering
    \begin{tabular}{lcccccc}
    \toprule
    {Method}\;\;   & {Mean$\pm$STD}\;\;& {Max}\;\; & {Min}\;\; & 5/5\;\;\\
    \midrule
    HOJ3D~\cite{ViewInvariantJoint3D}\;\;  & 63.55$\pm$5.23\;\; & 75.91 \;\;& 44.05 \;\;& 75.80\;\;\\    
    HOG3D~\cite{3Dgrad}\;\;  & 70.38$\pm$4.40\;\; & 82.78 \;\;& 55.26 \;\;& 82.78\;\;\\
    ROP~\cite{Wang2012} & - \;\;& -\;\; & -\;\; & 86.50\;\;\\
    Moving Pose~\cite{Zanfir} & - \;\;& -\;\; & -\;\; & \;\;$91.70^a$\;\;\\
    Actionlet~\cite{ActionLet2012}\;\;  & - \;\;& -\;\; & -\;\; & 88.20\;\;\\
    HON4D~\cite{HON4D}\;\;  & 81.88$\pm$4.45\;\; & 90.61\;\; & 69.31\;\; & 88.36\;\;\\
    DSTIP~\cite{DSTIP}\;\; & -\;\; & 89.30\;\; & -\;\; & -\;\;\\
    HDG~\cite{MyWACV14}\;\;  & 77.68$\pm$4.97\;\; & 86.13\;\; &  60.55\;\; & 83.70\;\;\\
    \midrule
    {\bf Holistic HOPC}\\
    constant $r$, constant $\tau$ \;  & 85.45$\pm$2.31\;\; & 92.39\;\; & 73.54\;\; & 91.64\;\;\\
      adaptive $r$, constant $\tau$ \;  & 84.78$\pm$2.89\;\; & 91.64\;\; & 72.41\;\; & 90.90\;\;\\
    constant $r$, adaptive $\tau$ \;  & 86.49$\pm$2.28\;\; & 92.39\;\; & 74.36\;\; & 91.64\;\;\\
     adaptive $r$, adaptive $\tau$ \;  & 85.01$\pm$2.44\;\; & 92.39\;\; & 72.94\;\; & 91.27\;\;\\
    \bottomrule
    \end{tabular}
  \label{tab:Action}
\end{table}

We also evaluated our local method with automatic spatial and temporal scale selection and achieved $90.90\%$ accuracy (subjects \{1,3,5,7,9\} used for training and the rest for testing). This is higher than $89.30\%$ of DSTIP~\cite{DSTIP} and $88.36\%$ of HON4D~\cite{HON4D}. Note that DSTIP~\cite{DSTIP} only reported the accuracy of the best fold and used additional steps such as mining discriminative features which can be applied to improve the accuracy of any descriptor. We did  not include such steps in our method.

\subsection{MSRGesture3D dataset}
The MSRGesture3D dataset~\cite{Wang2012} contains 12 American sign language gestures  performed 2-3 times by 10 subjects. For comparison with previous techniques, we use the leave-one-subject-out cross validation scheme proposed by~\cite{Wang2012}. Because of the absence of full body subjects (only hands are visible), we evaluate our methods in two settings only. Table \ref{tab:Gesture} compares our method to existing state-of-the-art methods excluding HOJ3D~\cite{ViewInvariantJoint3D} and Actionlet~\cite{ActionLet2012} since they require 3D joint positions which are not present in this dataset. Note that both variants of our method outperform all techniques by a significant margin achieving an average accuracy of $96.23\%$ which is $3.5\%$ higher than the nearest competitor HDG~\cite{MyWACV14}. We also tested our local method with automatic spatial and temporal scale selection and obtained an accuracy of $93.61\%$. 

\begin{table}[t] 
\centering \caption{\small Comparison with state-of-the-art methods on MSRGesture3D dataset}
    \begin{tabular}{lccccc}
    \toprule
    {Method}\;\;   & {Mean$\pm$STD}\;\;  & {Max}\;\; & {Min}\;\;\\
    \midrule
    \midrule
    HOG3D~\cite{3Dgrad} \;\; & 85.23$\pm$12.12\;\; & 100\;\; & 50.00\;\; \\    
    ROP~\cite{Wang2012}\;\; & 88.50\;\; & - & - \\
    HON4D~\cite{HON4D}\;\;  & 92.45$\pm$8.00\;\; & 100\;\; & 75\;\;\\
    HDG~\cite{MyWACV14}\;\;  & 92.76$\pm$8.80\;\; & 100 \;\;&  77.78\;\;\\    
    \midrule
    {\bf Holistic HOPC} \\
     adaptive $r$, constant $\tau$\;\;  & 95.29$\pm$6.24\;\; & 100\;\; & 83.67\;\;\\
     adaptive $r$, adaptive $\tau$\;\;  & 96.23$\pm$5.29\;\; & 100\;\; & 88.33\;\;\\
    
    \bottomrule
    \end{tabular}
  \label{tab:Gesture}
\end{table}
\subsection{ActionPairs3D dataset}
The ActionPairs3D dataset ~\cite{HON4D}  consists of depth sequences of six pairs of actions (Fig.~\ref{fig:DataSamples}) performed by 10 subjects. This dataset is challenging as each action pair has similar motion and shape. We used half of the subjects for training and the rest for testing as recommended by~\cite{HON4D} and repeated the experiments 252 folds. Table \ref{tab:Pairs} compares the proposed holistic HOPC descriptor in two settings with existing state-of-the-art methods. Our algorithms outperformed all techniques with $2.23\%$ improvement over the nearest competitor. Adaptive $\tau$ provides better improvement on this dataset compared to the previous two. We also evaluated our local method with automatic spatial and temporal scale selection and obtained $98.89\%$ accuracy using subjects \{6.7.8.9.10\} for training and the rest for testing.

\begin{table}[t]  
\centering \caption{\small Accuracy comparisons on the ActionPairs3D dataset. Mean$\pm$STD are computed over 252 folds. 5/5 means subjects \{6,7,8,9,10\} used for training and the rest for testing}
    \begin{tabular}{lcccccc}
    \toprule
    {Method}\;\;   & {Mean$\pm$STD}\;\;& {Max}\;\; & {Min}\;\; & 5/5\;\;\\
    \midrule
    HOJ3D~\cite{ViewInvariantJoint3D}\;\;  & 63.81$\pm$5.94\;\; & 67.22 \;\;& 50.56 \;\;& 66.67\;\;\\
    HOG3D~\cite{3Dgrad}\;\;  & 85.76$\pm$4.66\;\; & 85.56 \;\;& 65.00 \;\;& 82.78\;\;\\
    Actionlet~\cite{ActionLet2012}\;\;  & - \;\;& -\;\; & -\;\; & 82.22\;\;\\
    HON4D~\cite{HON4D}\;\;  & 96.00$\pm$1.74\;\; & 100\;\; & 91.11\;\; & 96.67\;\;\\
    \midrule
    {\bf Holistic HOPC}\\
    constant $r$, constant $\tau$\;\;  & 97.15$\pm$2.21\;\; & 100\;\; & 88.89\;\; & 97.22\;\;\\
    constant $r$, adaptive $\tau$\;\;  & 98.23$\pm$2.19\;\; & 100\;\; & 88.89\;\; & 98.33\;\;\\
    \bottomrule
    \end{tabular}
  \label{tab:Pairs}
\end{table}

\subsection{UWA3D Multiview Activity dataset}
\label{MyDataset}
We collected a new dataset using the Kinect to emphasize three factors: (1) Scale variations between subjects. (2) View-point variations. (3) All actions were performed in a continuous manner with no breaks or pauses. Thus, the start and end positions of body for the same actions are different. Our dataset  consists of 30 activities performed by 10 human subjects of varying scales: {\it one hand waving, one hand Punching, sitting down, standing up, holding chest, holding head, holding back, walking, turning around, drinking, bending, running, kicking, jumping, moping floor, sneezing, sitting down(chair), squatting, two hand waving, two hand punching, vibrating, falling down, irregular walking, lying down, phone answering, jumping jack, picking up, putting down, dancing,} and {\it coughing} (Fig.~\ref{fig:DataSamples}). To capture depth videos from front view, each subject performed two or three random permutations of the 30 activities in a continuous manner. For cross-view action recognition, 5 subjects performed  15 activities from 4 different side views (see Fig.~\ref{fig:Keypoints}-b). We organized our dataset by segmenting the continuous sequences. The dataset is challenging due to self-occlusions and high similarity. For example, {\it drinking} and {\it phone answering} actions have very similar motion and only the hand location in these actions is slightly different. As another example, {\it lying down} and {\it falling down} actions have very similar motion, but the speed of action execution is different. Moreover, some actions such as: {\it holding back, holding head,} and {\it answering phone} have self-occlusions. The videos were captured at 30 frames per second at a spatial resolution of $640 \times 480$. 

We evaluate our proposed methods in the same-view, and cross-view action recognition settings. The holistic approach is used to classify actions captured from the same view and the local approach is used for cross-view action recognition where the training videos are captured from front view and the test videos from side views.
\subsubsection{Same-view Action Recognition}
We selected half of the subjects as training and the rest as testing and evaluated our holistic method in two settings: (1) constant $r$, constant $\tau$, (2) constant $r$, adaptive $\tau$. Table \ref{tab:InHouse3Dsingleview} compares our methods with existing state of the art. Both variants of our algorithm outperform all methods achieving a maximum of $84.93\%$ accuracy. The adaptive $\tau$ provides minor improvement because there is no explicit action speed variation in the dataset.
To further test the robustness of our temporal scale selection  (adaptive $\tau$) to action speed variations we use depth videos of actions performed by half of the subjects captured at 30 frames per second as training data and depth videos of actions performed by the remaining subjects captured at 15 frames per second as test data. The average accuracy of our method using automatic temporal scale selection was $84.64\%$ which is higher than $81.92\%$ accuracy achieved by our method using constant temporal scale and the $76.43\%$ accuracy achieved by HON4D. Next, we swap the frame rates of the test and training data. The average accuracy of our method using automatic temporal scale selection was $84.70\%$ which is higher than $81.01\%$ accuracy achieved by our method using constant temporal scale. The accuracy of HON4D was $75.81\%$ in this case.

\begin{table}[t] 
\centering \caption{\small Accuracy comparison on the UWA3D Activity dataset for same-view action recognition}
    \begin{tabular}{lcccccc}
    \toprule
    {Method}\;\;   & {Mean$\pm$STD}\;\;& {Max}\;\; & {Min}\;\; \\
    \midrule
    HOJ3D~\cite{ViewInvariantJoint3D}\;\;  & 48.59$\pm$5.77\;\; & 58.70 \;\;& 28.93 \;\;\\
    HOG3D~\cite{3Dgrad}\;\;  & 70.09$\pm$4.40\;\; & 82.78 \;\;& 51.60 \;\;\\
    HON4D~\cite{HON4D}\;\;  & 79.28$\pm$2.68\;\; & 88.89\;\; & 70.14\;\; \\
    HDG~\cite{MyWACV14}\;\; & 75.54$\pm$3.64\;\; & 85.07\;\; &  61.90\;\; \\
    \midrule
    {\bf Holistic HOPC}\\
    constant $r$, constant $\tau$\;\;  & 83.77$\pm$3.09\;\; & 92.18\;\; & 74.67\;\;\\
    constant $r$, adaptive $\tau$\;\;  & 84.93$\pm$2.75\;\; & 93.11\;\; & 74.67\;\;\\
    \bottomrule
    \end{tabular}
  \label{tab:InHouse3Dsingleview}
\end{table}
\subsubsection{Cross-view Action Recognition}
In order to evaluate the STKP detector and HOPC descriptor for cross-view action recognition, we used front views of five subjects as training and side views  of the remaining five subjects as test. Table~\ref{tab:InHouse3Dcrossview} compares our method with existing state-of-the- art holistic and local methods for cross-view action recognition. Note that the performance of all other methods degrades when the subjects perform actions at different viewing angles. This is not surprising as existing methods assume that actions are observed from the same viewpoint i.e. frontal. For example, HON4D achieved $86.55\%$ accuracy when the training and test samples were in the same view (frontal). The average accuracy of HON4D dropped to $48.89\%$ when the training samples were captured from front view and the test samples were captured from four different side views. We also observed that the performance of existing methods did not degrade only for actions like {\it standing up, sitting down}, and {\it turning around}. This is due to the distinctness of these actions regardless of the viewpoint. 

We test two variants of our method. First, we apply our STKP detector on 3D pointcloud sequences and use the raw values of fitted hyper-surface as features. The average accuracy obtained over the four different side views ($\pm25^\circ$ and $\pm50^\circ$) was $76.56\%$ in this case. Next, we use the STKP detector combined with the proposed HOPC descriptor. This combination achieved the best average accuracy i.e. $82.23\%$. Comparison with other methods and the accuracy of each method on different side views are shown in Table~\ref{tab:InHouse3Dcrossview}. These experiments demonstrate that our STKP detector in conjunction with HOPC descriptor significantly outperforms state-of-the-art methods for cross-view as well as same-view action recognition. 

\begin{table}[t] 
\centering \caption{\small Cross-view action recognition  on the UWA3D Multiview Activity dataset. Depth sequences of five subjects at $0^o$ are used for training and the remaining subjects at $0^o$ and 4 different side-views are used for testing. Average accuracy is computed only for the cross-view scenario}
    \begin{tabular}{lcccccc}
    \toprule
    \;\;   & \;\;& \;\;& View angle\;\; & \;\; & \;\;& \;\;\\
    {Method}\;\;   & {$0^\circ$}\;\;& {$-25^\circ$}\;\;& {$+25^\circ$}\;\; & {$-50^\circ$}\;\; & {$+50^\circ$}\;\;& {Average}\;\;\\
    \midrule
    {\bf Holistic Methods}\\    
    HON4D~\cite{HON4D}\;\; & 86.55\;\; & 62.22\;\; & 60.00\;\;& 35.56\;\;&37.78\;\;&48.89\\
    HDG~\cite{MyWACV14}\;\; & 79.13\;\; & 60.00\;\; & 64.44\;\;& 33.33\;\;&35.56\;\;&48.33\\
    \midrule
    {\bf Local Methods}\\    
    HOJ3D~\cite{ViewInvariantJoint3D}\;\; & 63.34\;\; & 60.00\;\; & 62.22\;\;& 37.78\;\;&40.00\;\;&50.00\\
    DSTIP+DCSF~\cite{DSTIP}\;\; & 80.80\;\; & 66.67\;\; &  71.11\;\;&35.56\;\;&40.00\;\;&53.33\\
    \midrule      
    STKP+hyper-surface fitting\;\;  & 87.39\;\; & 81.33\;\; &  82.67\;\;& 71.11\;\;& 71.11\;\;& 76.56\\    
    STKP+HOPC\;\;  & {\bf 91.79}\;\; & {\bf 86.67}\;\; &  {\bf 88.89}\;\;&{\bf 75.56}\;\;&{\bf 77.78}\;\;&{\bf 82.23}\\
    \bottomrule
    \end{tabular}
  \label{tab:InHouse3Dcrossview}
\end{table}

\section{Conclusion}
Performance of current 3D action recognition techniques degrades in the presence of viewpoint variations across the test and the training data. We proposed a novel technique for action recognition which is more robust to action speed and viewpoint variations. A new descriptor, Histogram of Oriented Principal Components (HOPC), and a keypoint detector are presented. The proposed descriptor and detector were evaluated for activity recognition on three benchmark datasets.  We also introduced a new multiview public dataset and showed the robustness of our proposed method to viewpoint variations. 
\section*{Acknowledgment} 
This research was supported by ARC Discovery Grant DP110102399.

\bibliographystyle{splncs03}
\bibliography{BibECCV}
\end{document}